\documentclass[runningheads]{llncs}
\usepackage{graphicx}
\usepackage{multirow}
\usepackage{comment}
\usepackage{enumitem}
\usepackage[hidelinks]{hyperref}
\begin{document}

\title{Matching with Transformers in MELT} %

\author{Sven Hertling\inst{1}\thanks{The authors contributed equally to this paper.} \orcidID{0000-0003-0333-5888} \and 
Jan Portisch\inst{1,2}$^\star$ \orcidID{0000-0001-5420-0663} \and
Heiko Paulheim\inst{1}\orcidID{0000-0003-4386-8195}}

\authorrunning{Sven Hertling, Jan Portisch, and Heiko Paulheim}

\institute{Data and Web Science Group, University of Mannheim, Germany\\
	\email{\{sven, jan, heiko\}@informatik.uni-mannheim.de} \and
	SAP SE Business Technology Platform - One Domain Model, Walldorf, Germany\\
\email{\{jan.portisch\}@sap.com}	
}

\maketitle 
\begin{abstract}

One of the strongest signals for automated matching of ontologies and knowledge graphs are the textual descriptions of the concepts. The methods that are typically applied (such as character- or token-based comparisons) are relatively simple, and therefore do not capture the actual meaning of the texts. With the rise of transformer-based language models, text comparison based on meaning (rather than lexical features) is possible. 
In this paper, we model the ontology matching task as classification problem and present approaches based on transformer models.  We further provide an easy to use implementation in the MELT framework which is suited for ontology and knowledge graph matching.
We show that a transformer-based filter helps to choose the correct correspondences given a high-recall alignment and already achieves a good result with simple alignment post-processing methods. 

\end{abstract}
\section{Introduction}
\emph{Ontology Matching} is the non-trivial task of finding correspondences between classes, properties, and instances of two or more ontologies. The match operation can be seen as a function $f$ which returns an alignment $A$ given two ontologies $O_1$ and $O_2$: $f(O_1, O_2) = A$. The alignment is a set of correspondences in the form $\langle e_1, e_2, r \rangle$ where $e_1 \in O_1$, $e_2 \in O_2$, and $r$ is some relation which holds between the two concepts; in this paper $r$ is always equivalence ($\equiv$).

Multiple techniques exist to perform the matching operation in an automated manner~\cite{euzenat_ontology_2013_ch_4}. 
Labels and descriptions are one of the strongest signals concerning the semantics of an element of a knowledge graph. 
Here, matcher developers often borrow strategies from the natural language processing (NLP) community to determine similarity between two strings.

Since the attention mechanism~\cite{DBLP:conf/nips/VaswaniSPUJGKP17} has been presented, so called transformer models gained a lot of traction in the NLP area and transformer models achieved remarkable results on tasks such as machine translation~\cite{DBLP:conf/nips/VaswaniSPUJGKP17} or question answering~\cite{DBLP:conf/naacl/DevlinCLT19,DBLP:journals/corr/abs-1910-03771}.

In this paper, we bring transformers to the ontology matching task. Our contributions are twofold: Firstly, we present a transformer extension to the Matching and EvaLuation Toolkit (MELT), which allows users to easily exploit state-of-the-art pre-trained transformer models like BERT~\cite{DBLP:conf/naacl/DevlinCLT19} or RoBERTa~\cite{DBLP:journals/corr/abs-1907-11692} in their matching pipelines. Secondly, we evaluate different transformer-based matching approaches, and we discuss the strengths and weaknesses of transformer models in the matching domain.

\section{Related Work}

\emph{Transformers} are deep learning architectures which combine stacked encoder layers with a self-attention~\cite{DBLP:conf/nips/VaswaniSPUJGKP17} mechanism. These architectures are typically applied in unsupervised pre-training scenarios with massive amounts of data. Since transformers achieved very good results in the natural language processing (NLP) domain, they are also used in other domains. Brunner and Stockinger~\cite{DBLP:conf/edbt/BrunnerS20}, for instance, apply transformers for the task of entity matching and show that they achieve better results than classical deep learning models. Peeters et al.~\cite{DBLP:conf/vldb/PeetersBG20} report good results on the similar task of product record matching. In a similar spirit, the DITTO entity matching system consists of a complete architecture (including blocking and data augmentation for fine-tuning) for entity matching that is based on transformer models~\cite{DBLP:journals/jdiq/LiLSWHT21}. It is evaluated on the ER-Magellan benchmark and achieves good results.

Applications of transformers for the pure ontology matching task are less frequent compared to the entity matching domain.
Wu et al. \cite{daemon} created a Deep Attentional Embedded Ontology Matching (DAEOM) system which jointly encodes the textual description as well as the network structure. It contains negative sampling approaches as well as automatic adjustments of thresholds.  

\section{Matching with Transformers}
Since transformer models are language models, it is a hard requirement that the elements in the ontology have labels or descriptions. We propose to model the match operation as an unbalanced binary classification problem where the classifier receives a correspondence and predicts whether this correspondence is correct or not. Eventually, only correct correspondences are kept. The match operation can be (i) \emph{complete} or (ii) \emph{partial}.

In a \emph{complete} matching setting, each element $e_{1i} \in O_1$ respectively  $e_{2i} \in O_2$ needs a textual representation. The latter can be obtained, for instance, by concatenating the URI fragment and all annotation properties. The transformer model then classifies each element in the Cartesian product of the ontologies to be matched.

Since the set of comparisons grows quadratically for the complete matching case, and matching with transformers can be computationally intensive, it is also possible to use a candidate generator which reduces the total number of comparisons. This candidate generator can be regarded as matching system which returns an alignment $A_C$. In the \emph{partial} case, we generate textual representations only for candidates in the alignment ($c \in A_C$) and perform a classification operation only for the correspondences $c \in A_C$. Therefore, focus of the candidate generator should be recall since the generator determines the theoretically largest attainable recall score of the system, i.e., for the final alignment $A$, $A \subseteq A_C$ holds. This approach can also be seen as an \emph{matching repair} technique.

\section{MELT Transformer Extension}

\subsection{MELT}
\emph{MELT}\footnote{\url{https://github.com/dwslab/melt/}}~\cite{DBLP:conf/i-semantics/HertlingPP19} is a framework for ontology, instance, and knowledge graph matching. It provides functionality for matcher development, tuning, evaluation, and packaging. It supports both, HOBBIT and SEALS, two heavily used evaluation platforms in the ontology matching community. Since 2021, MELT also supports the new \emph{Web Interface}\footnote{\url{https://dwslab.github.io/melt/matcher-packaging/web}} format which was designed for the OAEI. 
The core parts of the framework are implemented in Java, but evaluation and packaging of matchers implemented in other languages is also supported. Via the MELT ML extension~\cite{DBLP:conf/semweb/HertlingPP20}, ML libraries developed in Python can also be used by Java components.  
Since 2020, MELT is the official framework recommendation by the OAEI and the MELT track repository is used to provide all track data required by SEALS. MELT is also capable of rendering Web dashboards for ontology matching results so that interested parties can analyze and compare matching results on the level of correspondences without any coding efforts~\cite{DBLP:conf/esws/PortischHP20}.

In this work, we extend the ML component of MELT so that transformer operations can be called directly from the Java code. Therefore, we use the \emph{Hugging Face transformers}\footnote{\url{https://github.com/huggingface/transformers}} library~\cite{wolf-etal-2020-transformers} which allows to use and fine-tune many transformer models.

\begin{figure}[t]
    \centering
    \includegraphics[width=\textwidth]{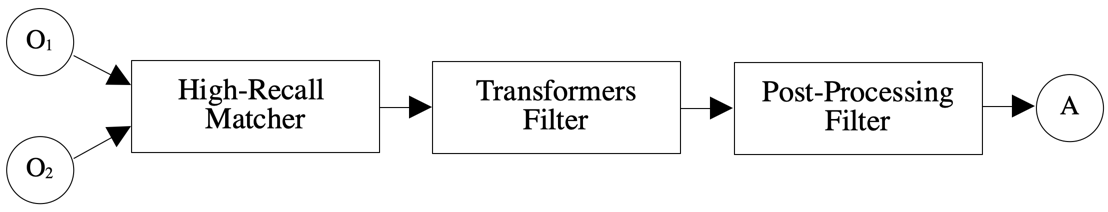}
    \caption{Recommended pipeline for the MELT transformer filter.}
    \label{fig:filter_architecture}
\end{figure}

\subsection{Obtaining Textual Descriptions from Resources}
In order to serialize textual descriptions, MELT offers various classes extending the \texttt{Text\-Extractor} interface. For any given resource, those return extracted text as a set of strings. They do not normalize the text because this is a post processing step. They merely select specific literals,  URI fragments, etc.
In our experiments, we use three of those extractors. They are ordered by the number of strings which are returned (most strings to fewest strings)\footnote{A more detailed overview can be found in the user guide:\\ \url{https://dwslab.github.io/melt/matcher-development/matching-with-jena\#textextractors}}:

\texttt{Text\-Extractor\-Set} returns the highest amount of literals because it retrieves all literals where the URI fragement of the property is either a label, name, comment, description, or abstract. This includes also \texttt{rdfs:\-label} and \texttt{rdfs:comment}. Furthermore, the properties \texttt{pref\-Label}, \texttt{alt\-Label}, and \texttt{hidden\-Label} from the \texttt{skos} vocabulary\footnote{\url{http://www.w3.org/TR/skos-reference}} are included, as well as the longest literal (based on the lexical representation of it). Additionally, all properties which are defined as \texttt{owl:Annotation\-Property} are followed in a recursive manner in case the object is not a label but a resource. In such a case, all annotation properties of this resource are added. The extractor reduces the potentially large set of literals by comparing the normalized texts and only returns the ones which are not identical (note here that the original literals are returned, not the normalized ones). 

The \texttt{Text\-Extractor\-Short\-And\-Long\-Texts} reduces the set of literals further by checking if a normalized literal is fully contained in another literal. In this case, the literal is not returned. This is only applied within the two groups of long and short texts to extract not only a long abstract but also a short label.
Label-like properties are regarded as short texts, while comment/description properties are regarded as long texts.

The \texttt{Text\-Extractor\-For\-Transformers} extracts the smallest number of literals (out of the text extractors presented here) by returning exclusively labels that are not contained in other labels (without distinguishing in long and short texts). This results in reducing the set of strings even more because labels which appear in a comment are also not returned.

\subsection{Transformers in the Matching Pipeline}
In order to allow for re-usable matching code, MELT allows to chain matchers to build a dedicated matching pipeline for various problems. In such a pipeline, each matcher receives the alignment of the previous component together with the ontologies that are to be matched (and optionally configuration parameters).

MELT differentiates between matchers and filters. A filter is a component which does not add new correspondences to the alignment but instead further processes the given alignment by (1) removing correspondences and/or (2) adding new confidence / feature weights to existing correspondences. 

\begin{figure}[t]
\centering
\includegraphics[width=0.8\textwidth]{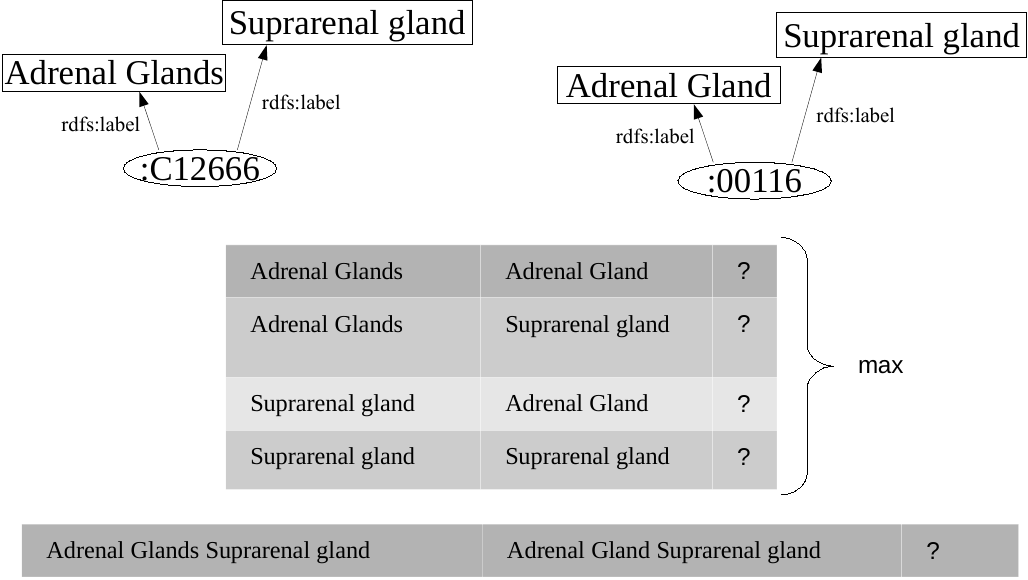}
\caption{Optional multi-text mechanisms implemented in class \texttt{Transformers\-Filter}.}
\label{fig:multitext}
\end{figure}

Since the transformer evaluation of the Cartesian product of descriptions is not a scalable option for most test cases, MELT offers the usage of transformers as a filter through class \texttt{Transformers\-Filter}. The training process is implemented using TensorFlow 
and PyTorch, 
the user can decide which implementation shall be used. Therefore, we recommend a transformer-based matching pipeline as shown in Figure~\ref{fig:filter_architecture}: In a first step, we use a matcher that generates a recall-oriented alignment. The transformer filter will then use the correspondences in the latter alignment to calculate the estimated similarity. The similarity is calculated by first serializing the textual descriptions of each correspondence to a CSV file. Textual descriptions are obtained by a \texttt{TextExtractor}. In case there are multiple textual descriptions available, two modes are implemented: (1) A multi-text option (depicted in Figure~\ref{fig:multitext}), which serializes all combinations of the individual texts; eventually, the maximum similarity will be used. (2) A single-text option which concatenates all textual elements.

After serializing the texts to be compared to a file, the ML Python server is started in the background and predicts the likelihood of a match given the textual description of each correspondence. It is optionally also possible to filter the alignment, for instance, by using a threshold or by reducing the alignment to a one-to-one alignment if applicable.

The MELT extension presented in this paper is publicly available in the main branch\footnote{\url{https://github.com/dwslab/melt/}} together with a reference implementation\footnote{\url{https://github.com/dwslab/melt/tree/master/examples/transformers}} that was used to run the experiments. The new features are documented in the MELT user guide\footnote{\url{https://dwslab.github.io/melt/}}.

\subsection{Generating Negatives}
In order to run a training process, such as fine-tuning a transformer, data is required for the training step. Positive correspondences can be obtained either from the reference\footnote{Note that convenience methods to do so exist in MELT such as\\ \texttt{generateTrackWithSampledReferenceAlignment(Track track, double fraction)} of class \texttt{TrackRepository}.} or from a high-precision matching system. However, negative examples are also required. Multiple strategies can be applied here. For example, negatives can be generated randomly using an absolute number of negatives (class \texttt{Add\-Negatives\-Randomly\-Absolute}) or a relative share of negatives  to be generated (class \texttt{Add\-Negatives\-Randomly\-Share}). If the gold standard is not known, it is also possible to exploit the one-to-one assumption and add random correspondences involving elements that already appear in the positive set of correspondences (class \texttt{Add\-Negatives\-Randomly\-One\-One\-Assumption}). The new extension to the MELT ML module contains multiple out-of-the box strategies that are already implemented as matching components which can be used within a matching pipeline. All of them implement the new interface \texttt{AddNegatives}. Since multiple flavors can be thought of (e.g. generating type homogeneous or type heterogeneous correspondences), a negatives generator can be easily written from scratch or customized for specific purposes. MELT offers some helper classes to do so such as \texttt{Random\-Sample\-OntModel} which can be used to sample elements from ontologies.

\begin{figure}[t]
    \centering
    \includegraphics[width=\textwidth]{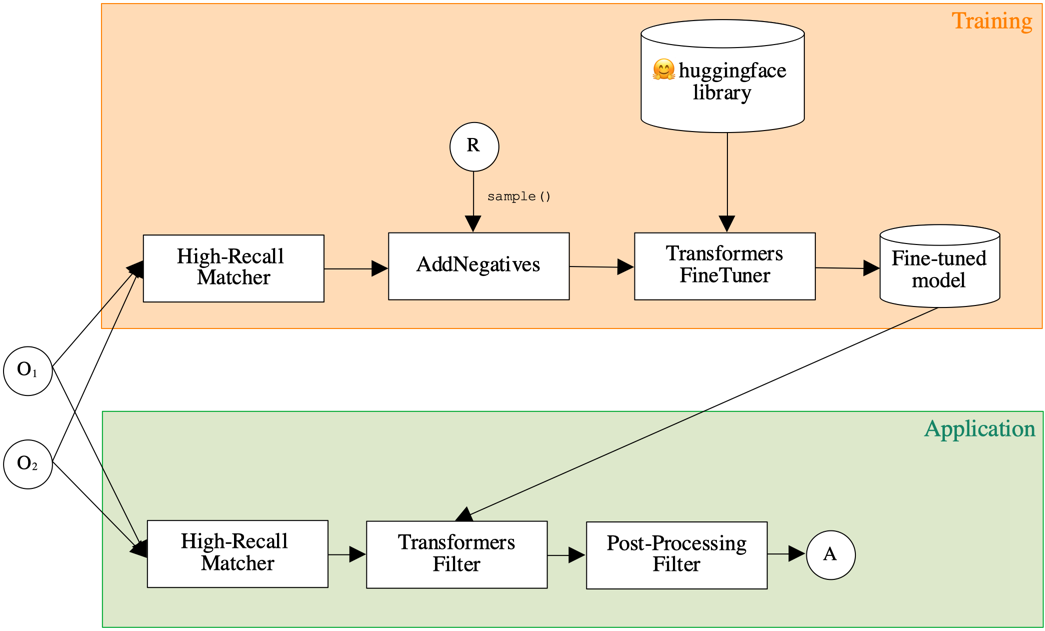}
    \caption{Proposed fine-tuning pipeline: The training step is represented by the components in the orange (upper) box, the application step of the fine-tuned model by the components in the green (lower) box. Note that the \emph{high-recall matcher} is identical in both steps.}
    \label{fig:fine_tune_architecture}
\end{figure}

Since the (partial) reference alignments of OAEI tasks are known and the one-to-one assumption holds, we propose to generate negatives using the same high-recall matcher that is also used in the matching pipeline and to apply the one-to-one sampling strategy: Given the reference and the alignment produced by some high-recall matcher, we determine the wrong correspondences as correspondences where only one element is found in the reference (but not the complete correspondence) and add them to the training set. This is implemented in class \texttt{AddNegativesViaMatcher}. Note that for this approach, the reference alignment does not have to be complete. One advantage here is that the characteristics of training and test set are very similar (such as the share of positives and negatives). This process is visualized in Figure~\ref{fig:fine_tune_architecture}.

\subsection{Fine-Tuning Transformers in MELT}
A transformer model can be used as is (particularly, if the application is equal or very similar to its training objective) or be fine-tuned for a specific task. The default transformer training objectives are not suitable for the task of ontology matching. 

Therefore, a pre-trained model needs to be fine-tuned. Once a training alignment is available, class \texttt{Transformers\-Fine\-Tuner} can be used to train and persist a model. Like the \texttt{Transformers\-Filter}, the \texttt{Transformers\-Fine\-Tuner} is a matching component that can be used in a matching  pipeline.\footnote{Note that this pipeline can only be used for training and model serialization. For the application of the model within a matching pipeline, \texttt{Transformers\-Filter} must be used.} Such a training pipeline is visualized in the orange (upper) part of Figure~\ref{fig:fine_tune_architecture}: A high-recall matcher can be used to generate candidates and negatives can be generated using a sampled reference (or a reference-like alignment). Repeated calls of the match method will extend the number of training candidates, the actual training is performed when calling method \texttt{finetuneModel}. This setup allows to train one model given multiple test cases. The implementation allows, for instance, to train a fine-tuned model per test case, per track, or a global model for multiple tracks. In this paper, we fine-tune the model per track to cover their individual characteristics.

\subsection{Hyperparameter Optimization}
\label{hyperparameter}
By default, the fine-tuning of the transformer models is executed with the standard training parameters such as a fixed number of epochs (3), a learning rate of $5 \cdot 10^{-5}$ etc. (those default values originate from the transformers library\footnote{\url{https://huggingface.co/transformers/main\_classes/trainer.html\#trainingarguments}}).
In hyperparameter optimization, a simple grid search is often applied.
But such a tuning method has some disadvantages: (1) each run (parameter combination) needs to be executed until the end to analyze the performance (2) all combinations need to be executed (no information about previous runs are taken into account). Bayesian Optimization~\cite{bayesian_optimization} solves the latter problem by modeling the performance based on the chosen hyperparameters. Thus, parameter combinations which do not look too optimistic are not tried out. Furthermore, runs can be canceled when the optimizing metric does not look promising.

Due to the fact that training of transformer based models is rather slow, even more sophisticated methods need to be applied. One of them is population based training~\cite{population_based_training} (PBT). 
Given a population of models, each is trained and evaluated after one epoch.
Some models trained with a given parameter combination perform better than others.
The better models are duplicated (via checkpointing of model weights) and replace the weaker models to keep the population size fixed. This step is called \emph{exploit} in PBT. Another step, called \emph{explore}, changes the hyperparameters during the training (e.g. the learning rate after the $2^{nd}$ epoch). With all these mechanisms, it is possible to explore a wide range of parameter in a shorter time frame.
PBT is implemented already in \emph{Ray Tune}~\cite{liaw2018tune} and uses distributions to describe the search space.
Furthermore, it is also used by the transformers library.
The initial hyperparameter search space looks as follows:
\begin{itemize}[topsep=0pt,itemsep=-1ex,partopsep=1ex,parsep=1ex]
    \item \texttt{learning rate}: loguniform distribution between $10^{-6}$ and $10^{-4}$
    \item \texttt{epochs}:  random choice between $1$ and $5$
    \item \texttt{seed}: uniform distribution between $1$ and $40$
    \item \texttt{batch size}: random choice of 4, 8, 16, 32, 64
\end{itemize}
The search space of the batch size is adjusted by the maximum possible values before the hyperparameter tuning starts. It will determine the maximum batch size by training for one step with the batch size of 4 and checking for out of memory errors. If this does not happen, the batch size will be increased in every step by multiplying the value by 2 (such that only powers of 2 are tried out). The final adjusted search space will be all powers of 2 starting from four until the maximum batch size is reached.

The seed is also optimized because different initializations of the classification head of the model can also improve the final metric. The reason behind this is that most models are trained on the masked language modeling task and need a classification layer (usually a linear layer on top of the pooled output) to create the final prediction. This linear layer is initialized with different random weights.

As described above, the hyperparameters can also be changed during training. The following parameters are mutated: \texttt{weight decay}: uniform distribution between 0.0 and 0.3; \texttt{learning rate}, and \texttt{batch size} as defined above.

The metric which is optimized can be chosen from the following KPIs: loss (of the model), accuracy, $F_1$, recall, precision, or AUC.
The last one is the default because in a later step in the matching pipeline, the confidence of a correspondence is important for filtering or selection. AUC optimizes this confidence such that all negatives have a low value and all positives a high one. Furthermore, it allows to decide which model is better even if they have the same F-measure.
The hyperparameter tuning can be easily performed in MELT with class \texttt{Transformers\-Fine\-Tuner\-Hp\-Search}. It has the same interface as the fine-tuning class but when calling the \texttt{finetuneModel} method, the hyperparameter search is started.

\section{Exemplary Analysis}

\subsection{Experiments}
In order to show the effectiveness of transformers for matching in MELT, we performed multiple experiments -- each focuses on a different aspect: (1) We evaluate an off-the-shelf transformer model in a zero-shot setting for three OAEI tracks: \emph{Anatomy}, \emph{Conference}, and \emph{Knowledge Graph}(KG)~\cite{DBLP:conf/semweb/HofmannPPHP17,DBLP:conf/esws/HertlingP20}, (2) we fine-tune well-known models and evaluate them with a sampling rate of 0.2 for the same tracks, (3) for the anatomy track and a fixed model, the sampling rates are modified and the performance is analyzed, (4) for the same track and model we optimize the hyperparameters and analyze their impacts. 

We use the following transformer models from the huggingface repository: \texttt{bert-base-cased}~\cite{DBLP:conf/naacl/DevlinCLT19}, 
\texttt{roberta-base}~\cite{DBLP:journals/corr/abs-1907-11692},  
and \texttt{albert-base-v2}~\cite{DBLP:conf/iclr/LanCGGSS20}. 
This sample is selected since these models are  well known and  often used according to the model hub\footnote{\url{https://huggingface.co/models}} of huggingface.

The matching pipeline consists of 4 components: (1) high-recall matcher, (2) transformer filter, (3) confidence threshold cut-off filter, and (4) max weight bipartite partitioning filter.

The high-recall matcher adds candidates with overlapping tokens, the transformer filter assigns a confidence to each candidate found in the previous step. 
An optimal threshold is determined to filter out non-matches. The threshold is calculated not with the complete gold standard but merely with the correspondences that were sampled for the training step. Therefore, the \texttt{Confidence\-Finder} class has been extended to work also with incomplete gold standards.
Lastly, the max weight bipartite partitioning filter enforces a one-to-one alignment.

\subsection{Results}
In the following, the results to all experiments are presented. The first part covers the zero-shot approach as well as the fine-tuning. Afterwards, we report on the impact of different sampling sizes and the results of the hyperparameter search.

\subsubsection{Zero-shot and Fine-tuning}
The results of the zero-shot and fine-tuning experiments are depicted in Table~\ref{tab:results}.
The \texttt{Simple\-String} baseline is a simple matcher which we use as a baseline. The high-recall matcher is the one which is used as a first step in the pipeline in the zero-shot as well as in the fine-tuning setup. 
This also means that the recall value of this matcher is automatically an upper bound for the recall because the transformer-based filtering will not add any new correspondences.
For the zero-shot case where an already fine-tuned model is applied directly (in this case no reference sampling is necessary), we selected a dataset which is rather close to our setup. Due to the fact that paraphrasing is very similar to the task of finding same concepts, the Microsoft Research Paraphrase Corpus~\cite{mrpc_dataset} is selected. The bert-base-cased model already exists in the hugginface hub and is fine-tuned on this dataset. It performs best on the conference track but these results should be taken with care because of the small amount of correspondences and textual descriptions in this track. For the anatomy and knowledge graph track, the fine-tuned models perform much better. For the former dataset, \texttt{albert} outperformed \texttt{bert} and \texttt{roberta} by a large margin. In the KG track, \texttt{bert} performed much better.
One reason why different models perform better is the different characteristics of the labels and comments.

For conference and anatomy, the \texttt{Text\-Extractor\-Set} is used with the multitext setup to generate many classification examples whereas for the KG track the \texttt{Text\-Extractor\-For\-Transformers} is used to extract less literals which are then concatenated together to create only one classification example for each correspondence.

\begin{table}[t]
\begin{tabular}{|c|l|l|l|l|l|l|l|l|l|l|}
\hline
\multicolumn{2}{|l|}{\multirow{2}{*}{}} & \multicolumn{3}{c|}{\textbf{Conference}} & \multicolumn{3}{c|}{\textbf{Anatomy}} & \multicolumn{3}{c|}{\textbf{\begin{tabular}[c]{@{}c@{}}Knowledge\\ Graph\end{tabular}}} \\ \cline{3-11} 
\multicolumn{2}{|l|}{} & \multicolumn{1}{c|}{\textbf{P}} & \multicolumn{1}{c|}{\textbf{R}} & \multicolumn{1}{c|}{\textbf{F1}} & \multicolumn{1}{c|}{\textbf{P}} & \multicolumn{1}{c|}{\textbf{R}} & \multicolumn{1}{c|}{\textbf{F1}} & \multicolumn{1}{c|}{\textbf{P}} & \multicolumn{1}{c|}{\textbf{R}} & \multicolumn{1}{c|}{\textbf{F1}} \\ \hline
\multirow{2}{*}{\textbf{Baseline}} & \textbf{SimpleString} & 0.710 & 0.498 & 0.586 & 0.964 & 0.708 & 0.816 & 0.909 & 0.727 & 0.808 \\ \cline{2-11} 
 & \textbf{High Recall} & 0.450 & 0.561 & 0.179 & 0.037 & 0.942 & 0.071 & 0.167 & 0.915 & 0.283 \\ \hline
\multicolumn{1}{|l|}{\textbf{Zero-Shot}} & \textbf{\begin{tabular}[c]{@{}l@{}}bert-base-cased\\ (mrpc-tuned)\end{tabular}} & 0.650 & 0.548 & \textbf{0.594} & 0.531 & 0.817 & 0.644 & 0.739 & 0.714 & 0.726 \\ \hline
\multirow{3}{*}{\textbf{\begin{tabular}[c]{@{}c@{}}Fine-Tuned\\ (per Track)\end{tabular}}} & \textbf{bert-base-cased} & 0.748 & 0.361 & 0.487 & 0.726 & 0.689 & 0.707 & 0.941 & 0.789 & \textbf{0.859} \\ \cline{2-11} 
 & \textbf{roberta-base} & 0.667 & 0.498 & 0.570 & 0.715 & 0.749 & 0.732 & 0.400 & 0.388 & 0.393 \\ \cline{2-11} 
 & \textbf{albert-base-v2} & 0.812 & 0.397 & 0.533 & 0.854 & 0.825 & \textbf{0.839} & 0.687 & 0.665 & 0.676 \\ \hline
\end{tabular}
\caption{Results of non-fine-tuned and fine-tuned transformer models (multi-text) with 20\% sampling from the reference alignment. As per OAEI customs, we report micro average scores for the conference and macro average scores for the KG track.}
\label{tab:results}
\end{table}

\normalsize

\subsubsection{Sampling Rates} 
\begin{figure}[t]
\centering
\includegraphics[scale=0.32]{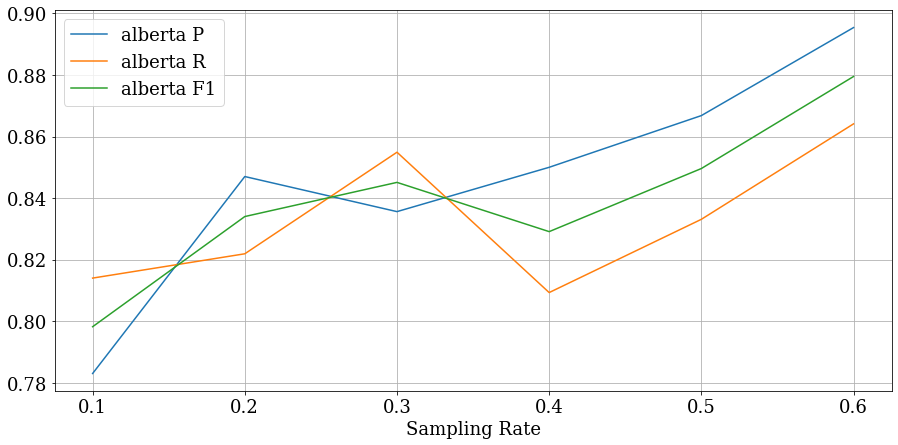}
\caption{\texttt{albert-base-v2} performance on the anatomy track using different reference sampling rates.}
\label{fig:sampling}
\end{figure}

We analyzed the performance of the best model on anatomy (albert) using varying sampling rates $s \in [0.1, 0.2, 0.3, 0.4, 0.5, 0.6]$ from the reference. The results are presented in Figure~\ref{fig:sampling}. Interestingly, fairly good performance can be achieved with very low sampling rates (10\% and 20\% respectively). Intuitively, the overall performance tends to increase with an increasing share of samples from the reference.

\subsubsection{Hyperparameter Tuning}
The hyperparameter tuning was executed for the anatomy track and the \texttt{albert\--base\--v2} model. The given search space in Section~\ref{hyperparameter} is used and overall 12 trials are sampled from it which is also the amount of the model population. The search needs 45 minutes to run in parallel on 4 GPUs (NVIDIA GeForce GTX 1080 Ti).
All other settings are the same as in the normal fine-tuning setup (thus, the numbers are comparable). With PBT, the precision could be improved by 0.02 to 0.874 whereas the recall is only a bit higher (0.832). In terms of F-Measure, the hyperparameter tuning additionally gives an improvement of 0.013 (eventually leading to an $F_1$ of 0.852).

\section{Conclusion and Outlook}
In this paper, we introduced a new matching component to the MELT framework which is based on transformer models. It allows to extract a textual description of the resource with so called text extractors and provides an easy option to apply and fine-tune transformer based models. We propose and evaluate an exemplary matching pipeline for transformer training and application. We hope that our implementation benefits the ontology matching community and enables other researchers to further explore this topic. 

In addition, we performed four experiments which demonstrate the capabilities of the newly implemented component. We showed that a transformer-based filter can improve a given alignment by providing a confidence for each correspondence based on its textual description. 
Moreover, we presented a sophisticated approach for hyperparameter tuning and showed that improvements can be achieved when optimizing the model hyperparameters.

Since the fine-tuning obviously has a large impact on the results, we will conduct further experiments on that step in the future. Examples include fine-tuning with text corpora from the domain of matching (e.g., biomedical texts for the anatomy track), or transfer learning setups where fine-tuning is conducted based on matching gold standards from other domains. 

Moreover, we plan to extend the implementation to also cover components that do not require any input alignment. These would also include matches which would not be possible with string comparison based systems. The library Sentence Transformers~\cite{reimers-2019-sentence-bert} allows to embed the textual description of a resource in such a way that similar entities are close in an embedding space. Thus, a search would be easily possible and would help in finding correspondences which might not share a lot of tokens but a similar meaning.

\subsubsection*{Acknowledgements}
The authors acknowledge support by the state of Baden-Württemberg through bwHPC.

%
%
\bibliographystyle{splncs04}
\bibliography{references}
\end{document}